\pdfoutput=1

\documentclass[11pt]{article}
\usepackage{graphicx}
\usepackage{booktabs}
\usepackage{subcaption}
\usepackage[preprint]{acl}

\usepackage{times}
\usepackage{latexsym}

\usepackage[T1]{fontenc}

\usepackage[utf8]{inputenc}

\usepackage{microtype}

\usepackage{inconsolata}

\title{How Much Annotation is Needed \\ to Compare Summarization Models?}

\newcommand*\samethanks[1][\value{footnote}]{\footnotemark[#1]}

\author{
\quad\textbf{Chantal Shaib$^1$\thanks{Work completed while at Adobe Research.}}\quad\quad
\textbf{Joe Barrow$^3$\samethanks}\quad\quad
\textbf{Alexa F. Siu$^2$}\quad\quad\\
\textbf{Byron C. Wallace$^1$}\quad\quad
\textbf{Ani Nenkova$^2$}\\
$^1$Northeastern University, $^2$Adobe Research, $^3$Pattern Data\\
\small\texttt{\{shaib.c, b.wallace\}@northeastern.edu}\\
\small\texttt{\{asiu, nenkova\}@adobe.com} \\
\small\texttt{joe.barrow@patterndataworks.com}\\
}
\begin{document}
\maketitle
\begin{abstract}
Modern instruction-tuned models have become highly capable in text generation tasks such as summarization, and are expected to be released at a steady pace. 
In practice one may now wish to choose confidently, but with minimal effort, the best performing summarization model when applied to a new domain or purpose. 
In this work, we empirically investigate the test sample size necessary to select a preferred model in the context of news summarization. 
Empirical results reveal that comparative evaluation converges quickly for both automatic and human evaluation, with clear preferences for a system emerging from under 100 examples. The human preference data allows us to quantify how well automatic scores can reproduce preference rankings across a variety of downstream summarization tasks. We find that, while automatic metrics are stable at smaller sample sizes, only some automatic metrics are able to moderately predict model win rates according to human preference. 
\end{abstract}

\section{Introduction} 
Instruction fine-tuned language models are highly capable summarizers, and new such models are now released often.
Continuously comparing such models using large, reference-based benchmark assessments is a costly task, especially if one wants to use them in a new domain. 
Here we demonstrate on data for news summarization that, in both human and automatic evaluations, preferences toward a summarization model emerge over test sets of about 50 samples. Collecting human judgements, GPT evaluations or---if available---human references for this size of dataset is reasonable.
Further, we validate GPT evaluations and two popular reference-based evaluations, ROUGE-1 and BERTScore, in terms of their ability to predict human preferences on a set of 36 testing contexts.
We collect human judgements in the context of three different summarization tasks and three sources of input. For these variations, we compute the accuracy of automated scores to reproduce human preferences between pairs of systems.

\section{Background}
Our goal is to establish the amount of test data needed to decide which of two summarization models produces better summaries, for a given distribution of the inputs (different sources of text to be summarized) and different task contexts for which the summary is to be used. 
\begin{figure*}
    \centering
    \includegraphics[width=\textwidth]{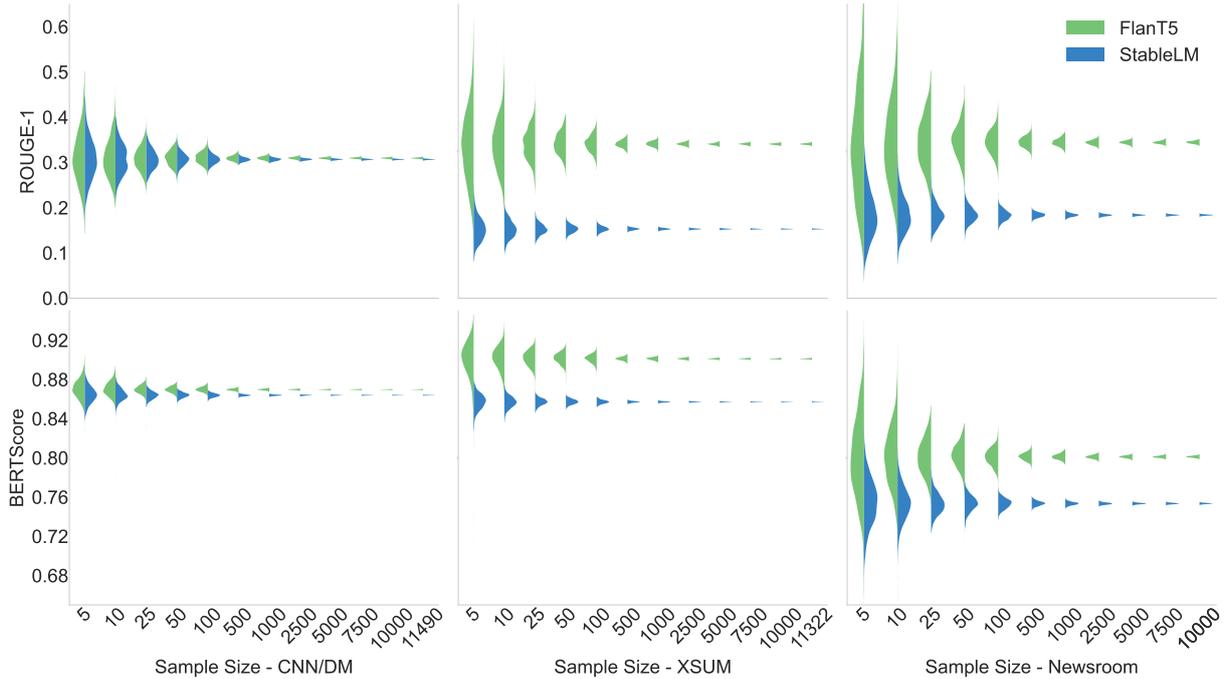}
    \caption{Distributions of average ROUGE-1 and BERTScores across 1000 re-samples. Differences between systems emerge clearly and quickly for XSUM and Newsroom.} 
    \label{fig:full_plots}
\end{figure*}

It is common to approach evaluation as a rate-then-compare task, in which outputs from systems are rated for quality on a scale, and then average scores are used to compare systems. 
But it is well known that inputs may differ considerably in difficulty ~\cite{nenkova-louis-2008-summarize}. 
Paired tests for statistical significance, 
that evaluate the differences of scores between two systems on the same input is the basis for comparison are therefore more appropriate \cite{rankel-etal-2011-ranking,dror-etal-2018-hitchhikers}. 
Most contemporary work has fully embraced this approach, largely abandoning scoring of outputs and instead soliciting preferences among two or more choices \cite{DBLP:conf/naacl/NovikovaDR18}. 
Given the developments in LLMs, pairwise win rates have become the \emph{de facto} standard for reporting comparisons between instruction fine-tuned models. 
In this work, we similarly adopt the win rate approach to comparing systems, and empirically identify the smallest test set size that reliably reveals preferences. 

Most closely related to our work is the study on estimating power of tests for statistical significance, i.e., the minimum test size necessary to detect statistical differences of a given size \cite{10.18653/v1/2020.emnlp-main.745}. Our work is aligned with the main question in this prior work, but we present empirical estimates of differences between systems, without making any assumptions of tests to be used or size of effect we want to detect.
Our findings can inform future work on power estimation.

Prior related work proposes ways of carrying out the evaluation, either automatically or manually \cite{laban2022summac, bert-score, fabbri-etal-2022-qafacteval, zhong-etal-2022-towards,10.48550/arxiv.2212.07981}, and of measuring the correlation between system rankings produced by human and automatic evaluations on a given benchmark~\cite{10.1613/jair.1.13715}. 
We do not propose new ways for evaluation but introduce a new method of validating automatic evaluations that does not rely on the benchmark, but rather measures the accuracy of automatic scores in reproducing human judgements across different input distributions and intended use-cases.

\section{Unnecessarily Large Benchmarks}
\label{sec:large_benchmarks}
We first compare two models, FlanT5-XXL \citep{https://doi.org/10.48550/arxiv.2210.11416} and StableLM \citep{gpt-neox-library} via automatic scores over three news summarization benchmarks: CNN/DM \citep{see-etal-2017-get, 10.5555/2969239.2969428}, XSUM \citep{Narayan2018DontGM}, and Newsroom \cite{N18-1065}. We use the test set splits of these datasets from Huggingface.\footnote{\url{https://huggingface.co/docs/datasets/index}} 

CNN/Daily Mail and XSUM contain about 10K test inputs.
The Newsroom test set split has over 100k samples. For efficiency, we randomly sample 10k examples from this set to scale it down to a size comparable to the other two datasets. 
We then generate summaries with FlanT5 and StableLM for all articles in the test sets, using the summarization prompts that these models have been trained on (see Appendix~\ref{appendix:prompts}).
For each test split we sample 1000 times with replacement smaller test set sizes ranging from $[5, \texttt{len}(dataset)]$. 
We evaluate the two models with the commonly used ROUGE-1 \citep{lin-2004-rouge} and BERTScore \citep{bert-score}.\footnote{We also run experiments with BLEU \citep{papineni-etal-2002-bleu} and SummaC-ZS \citep{10.1162/tacl_a_00453}, and report these results in Appendix \ref{appendix:addl_violin_plots}.}
Both scores compare a summary with a human-written reference summary. 
ROUGE does so using tokens, while BERTScore relies on embeddings.

We show score variations for FlanT5 and StableLM across the three datasets in Figure \ref{fig:full_plots}. 

For all three datasets, a preference for one of the models emerges early: The winning model as scored over 10k test points emerges after just 25-50 samples.\footnote{Even with respect to automatic evaluations, these findings have considerable implications. 
For many LLMs, simply producing outputs for all 10K test set is computationally expensive and slow. 
Manual (human) evaluation with such large test sets is practically impossible.}

Given these findings, we collect human judgements on 100 samples from each of the data sources, varying the task context in which the judgement is made. We also add GPT-4 as another summarization model to be evaluated, and later report the accuracy of GPT-based evaluation against the aggregated human judgements.

\section{Human Preferences}  

We hire annotators on Upwork\footnote{https://www.upwork.com/nx/enterprise-homepage/}. 
Specifically, we hire three individuals for CNN/DM and Newsroom, and one for XSUM. 
We select 100 inputs for annotation from each dataset, which given the trends we observed in the previous section, would be sufficient to reveal human preference.\footnote{See Appendix~\ref{appendix:annotation_details} for details about cost and hours for all annotations.}

\begin{figure*}
    \centering
    \includegraphics[width=\textwidth]{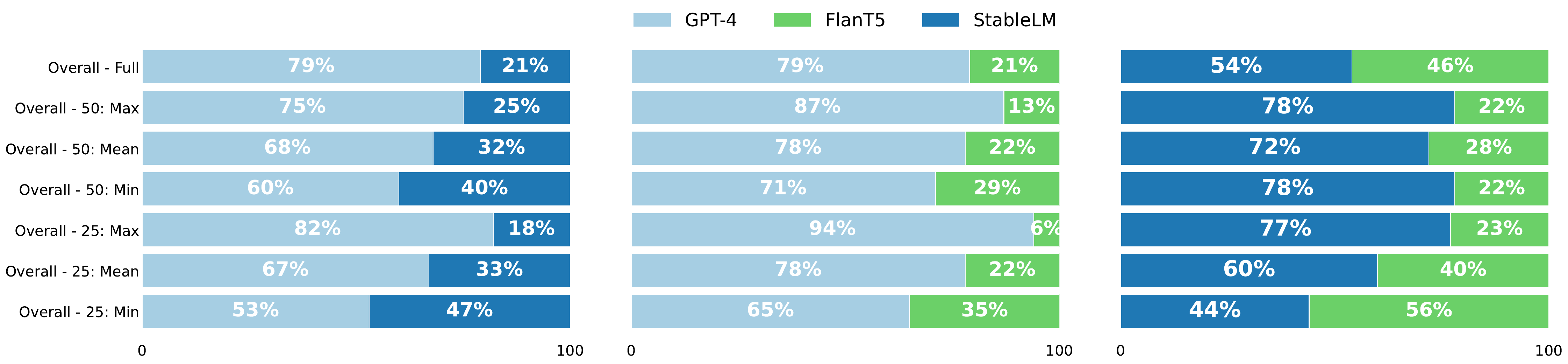}
    \caption{Aggregated annotator win rates for the CNN/DM dataset for the overall metric. Model preferences remain fairly stable across all sample sizes except in one case for sample size of 25.}
    \label{fig:win_rates_cnn}
\end{figure*}

We also add summaries produced by GPT-4 for evaluation on the smaller dataset. FlanT5, StableLM and GPT-4 represent encoder-decoder, 
decoder-only (open-source), and decoder-only (closed-source) models, respectively. 

We instruct annotators to rank the summaries for each input in order of preference. This is a typical evaluation setting in which win rates---the percentage of input for which the model was preferred over the other---provide the clearest score for each model pair. 

We provide three different scenarios to measure how preference may change based on context: \emph{(i)} Rank the summaries in order of preference; \emph{(ii)} Assuming you are monitoring the news for important world events, rank the summaries in order of preference; \emph{(iii)} Which summary best captures the main details of the event being reported on? \emph{(iv)} Which summary contains the fewest unnecessary details?\footnote{We also ask if the summaries have text quality issues (e.g., formatting, grammar, unusual symbols, or other artifacts). We then present the reader with the full article and ask them to mark if any of the summaries contain factual errors. We provide a brief analysis of these results in Appendix \ref{appendix:text_quality_factuality}.}

For GPT-4, we linearly append the summaries with the instructions and provide these as prompts to the model.  

\subsection{Stability of Preference}
First, we look at confirming whether smaller test samples are sufficient to make the same conclusion as with a larger sample. We apply the same procedure described in Section 3, where we resample 1000 test sets of size 25 and 50 from the 100 for which we have human judgements. Figure \ref{fig:win_rates_cnn} shows the win rates for the CNN/Daily Mail test set for each of the three pairs of models, on the full test set of 100 samples, as well as the min, max and average win rate recorded across the 1000 smaller test sets. 

While there is some variation in the strength of the preference for a model, the overall preference is preserved in the smaller samples. In only one case---the comparison between FlanT5 and StableLM---does the overall preference change for the minimum value of win rates from the one thousand samples  of size 25. With 50 samples in the evaluation set, all three of the minimum, maximum and average win rates lead to the same conclusion about which system in the pair is better as that from the full 100 sample test set. 

Similarly for the other two datasets, Newsroom and XSUM, none of the overall preferences flip for test sets of size 50 and only one minimum value for the 25 samples flips the preference. We provide the complete tables in Appendix \ref{appendix:human_eval_win_rates}.

These results indicate that, even for human evaluation, smaller test set samples ($n$=50) are adequate to conclude which is the preferred summarization model.

In many cases, the strength of the preference is of interest. As shown in the variation between the minimum and maximum win rates, the strength as captured by win rates can vary considerably depending on the test set. 
We leave for future work analysis of the test size required to obtain reliable conclusions about the strength of the preference. 

\subsection{Human Preference Varies by Task and Input Source}

We now turn to comparing model preferences relative to downstream task use. 

\begin{figure}
    \centering
    \includegraphics[width=0.5\textwidth]{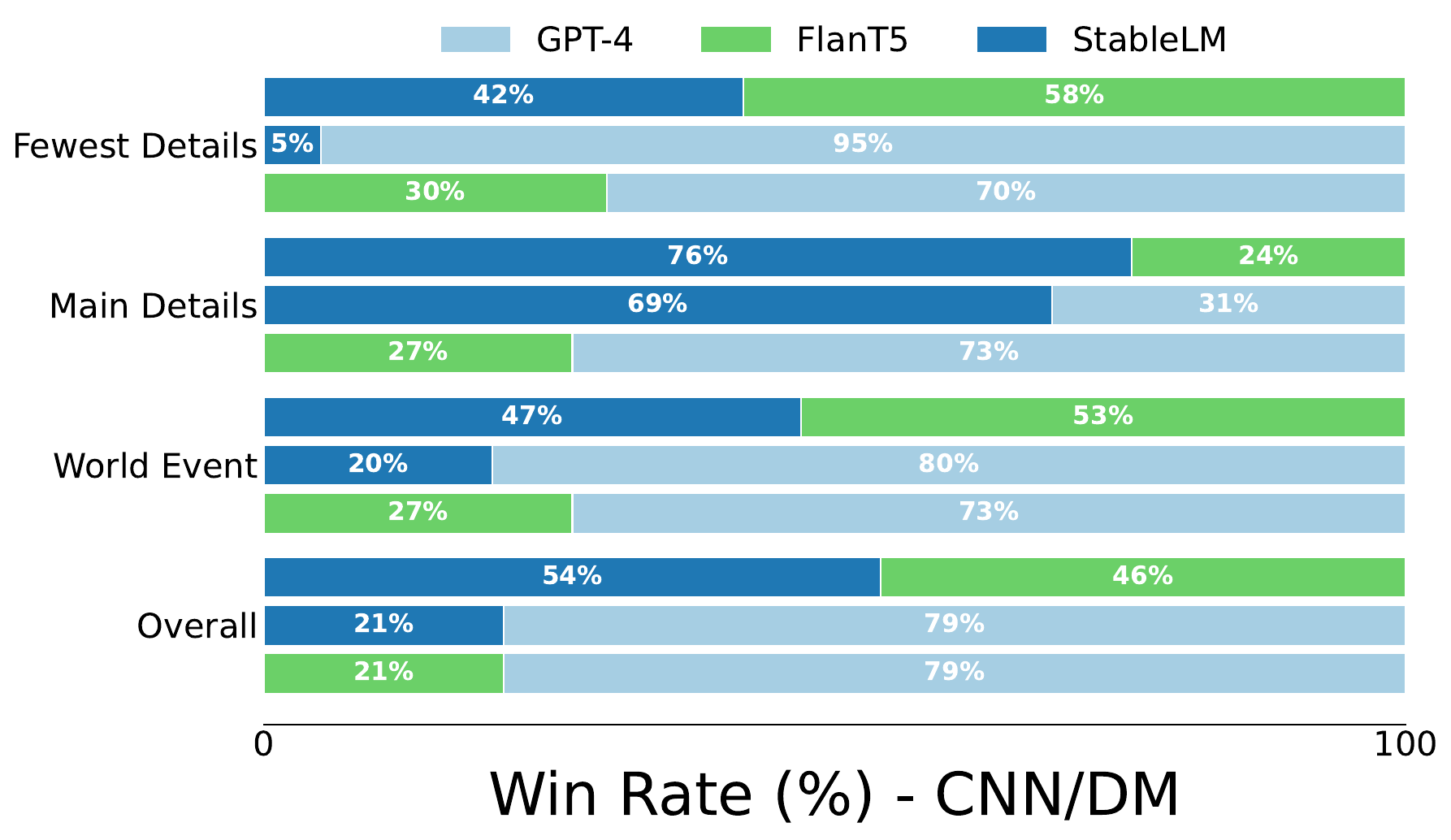}
    \caption{Aggregated annotator win rates across all metrics. Model preferences can change depending on the task setting.}
    \label{fig:win_rates_cnn_all_metrics}
\end{figure}

Figure \ref{fig:win_rates_cnn_all_metrics} shows the variation of aggregated preferences on the full 100 sample test set for CNN/Daily Mail. The context of the task can dramatically change the win rates for a given model.  When contextualized in a specific use-case, human preferences flip from the overall rating for two out of the three model comparisons. 

The overall win rate for StableLM over FlanT5 is 54\% indicating a weak preference for StableLM. In the world event use case however, the win-rate for FlanT5 increases to  53\%, flipping to a preference for FlanT5. Similarly, the win rate of StableLM over GPT-4 in the overall condition is 21\% but it flips to 76\% in the main details setting. The win rates of FlanT5 over GPT-4 remain stable across all tasks, always in favor of GPT-4.

Similarly, win rates according to the aggregate human preference for two systems vary depending with the source of data. 
In the next section we discuss how this observed variability changes the approach to validation of automatic evaluations.

\section{Validating Automatic Evaluation}
 We presented qualitative evidence that the context in which preferences are made change the human preferences dramatically. 
 We also provided clear examples of cases when human preference for the same two models can flip depending on the context. 
 This judgement variability poses a novel requirement for validating automatic evaluation approaches. 
 We cannot combine win rates across settings and compute correlations between human preferences and automatic scores because these come from different distributions. We do, however, have a sufficient number of pairs for comparison: 3 models evaluated on 3 sources of data, on 4 context of use. 
 This yields 9 overall preferences and 27 contextually dependent preferences. 

For four automatic methods for evaluation, we compute the accuracy of the automatic score in reproducing human preferences. Specifically, we compute the percentage of pairwise comparisons for which the automatic evaluation agrees with the human win rates on which system is the better one.
This is a coarse requirement because it does not capture the size of the win rate. For example the win rate of one system over another in human preferences is 51\% but an automatic score predicts that its win rate is 79\%, the automatic score will be considered accurate. 

Table \ref{tab:validation} shows the accuracy for four automatic evaluations: ROUGE-1, BERTScore, G-Eval, and GPT-4 as an annotator. In the case of GPT-4 as an annotator, we provide GPT-4 with the exact same instructions as the human annotators. For the first three approaches, a win for a model is declared if the score assigned by the method for this input is higher than that for the other model. In cases when the scores for an input are the same, there is a tie. In the fourth case, using GPT-4 as an annotator provides ratings, so the wins are decided by the ranking returned by GPT-4 (rather than a proxy score). 
In this case, there are no ties because the annotators were asked to do a forced choice comparison. We find that ROUGE-1 and GPT-4 as an annotator are able to moderately predict the aggregated human preferences across the different tasks, compared to BERTScore and G-Eval which are not able to do so as reliably.

\begin{table}[]
\centering
\begin{tabular}{@{}ll@{}}
\toprule
\textbf{Metric}      & \textbf{Accuracy (\%)}  \\ \midrule
ROUGE-1              & 78                      \\
BERTScore            & 56                      \\
G-Eval               & 44                      \\ 
GPT-4 (as annotator) & 78                  
\end{tabular}
\caption{Accuracy of automatic metrics compared to human evaluations. GPT-4 as-an-annotator and ROUGE-1 score have the highest accuracy in predicting which model is selected by human annotators in each setting task setting.}
\label{tab:validation}
\end{table}

\section{Conclusions}
We presented automatic and human evaluation designed to establish the minimum amount of data necessary to evaluate contemporary summarization models. Comparative evaluations establish which model performs better with test set of 50 inputs. For human evaluation, a test size of 50 is sufficient to confidently establish which is the model that people prefer. Human preference varies, however, depending on the intended use of the summary and on the source of data for summarization. This variation calls for new methods for validating automatic scores and we propose one. We find that all four automatic evaluations better than deciding preferences randomly but lead to erroneous conclusions for many pairwise comparisons.

\section*{Limitations}
We only evaluate over benchmark news datasets, where it is possible that our observations may not be reflected in other, more niche domains. In part, this choice is due to lack of availability of quality summarization datasets with references (and further motivating the need for evaluation over small samples), however it is important for future work to consider more specialized cases. Another limitation is that we do not collect human annotations nor GPT-4 summaries over the entire test set splits. This poses a challenge as collecting these evaluations and summaries over such a big dataset is costly. 

\section*{Acknowledgements}
We gratefully acknowledge the National Science Foundation (RI 2211954) for supporting this work. 

\bibliography{anthology,custom}

\begin{thebibliography}{20}
\expandafter\ifx\csname natexlab\endcsname\relax\def\natexlab#1{#1}\fi

\bibitem[{Andonian et~al.(2021)Andonian, Anthony, Biderman, Black, Gali, Gao, Hallahan, Levy-Kramer, Leahy, Nestler, Parker, Pieler, Purohit, Songz, Phil, and Weinbach}]{gpt-neox-library}
Alex Andonian, Quentin Anthony, Stella Biderman, Sid Black, Preetham Gali, Leo Gao, Eric Hallahan, Josh Levy-Kramer, Connor Leahy, Lucas Nestler, Kip Parker, Michael Pieler, Shivanshu Purohit, Tri Songz, Wang Phil, and Samuel Weinbach. 2021.
\newblock \href {https://doi.org/10.5281/zenodo.5879544} {{GPT-NeoX: Large Scale Autoregressive Language Modeling in PyTorch}}.

\bibitem[{Card et~al.(2020)Card, Henderson, Khandelwal, Jia, Mahowald, and Jurafsky}]{10.18653/v1/2020.emnlp-main.745}
Dallas Card, Peter Henderson, Urvashi Khandelwal, Robin Jia, Kyle Mahowald, and Dan Jurafsky. 2020.
\newblock \href {https://doi.org/10.18653/v1/2020.emnlp-main.745} {{With Little Power Comes Great Responsibility}}.
\newblock \emph{Proceedings of the 2020 Conference on Empirical Methods in Natural Language Processing (EMNLP)}, pages 9263--9274.

\bibitem[{Chung et~al.(2022)Chung, Hou, Longpre, Zoph, Tay, Fedus, Li, Wang, Dehghani, Brahma, Webson, Gu, Dai, Suzgun, Chen, Chowdhery, Narang, Mishra, Yu, Zhao, Huang, Dai, Yu, Petrov, Chi, Dean, Devlin, Roberts, Zhou, Le, and Wei}]{https://doi.org/10.48550/arxiv.2210.11416}
Hyung~Won Chung, Le~Hou, Shayne Longpre, Barret Zoph, Yi~Tay, William Fedus, Eric Li, Xuezhi Wang, Mostafa Dehghani, Siddhartha Brahma, Albert Webson, Shixiang~Shane Gu, Zhuyun Dai, Mirac Suzgun, Xinyun Chen, Aakanksha Chowdhery, Sharan Narang, Gaurav Mishra, Adams Yu, Vincent Zhao, Yanping Huang, Andrew Dai, Hongkun Yu, Slav Petrov, Ed~H. Chi, Jeff Dean, Jacob Devlin, Adam Roberts, Denny Zhou, Quoc~V. Le, and Jason Wei. 2022.
\newblock \href {https://doi.org/10.48550/ARXIV.2210.11416} {Scaling instruction-finetuned language models}.

\bibitem[{Dror et~al.(2018)Dror, Baumer, Shlomov, and Reichart}]{dror-etal-2018-hitchhikers}
Rotem Dror, Gili Baumer, Segev Shlomov, and Roi Reichart. 2018.
\newblock \href {https://doi.org/10.18653/v1/P18-1128} {The hitchhiker{'}s guide to testing statistical significance in natural language processing}.
\newblock In \emph{Proceedings of the 56th Annual Meeting of the Association for Computational Linguistics (Volume 1: Long Papers)}, pages 1383--1392, Melbourne, Australia. Association for Computational Linguistics.

\bibitem[{Fabbri et~al.(2022)Fabbri, Wu, Liu, and Xiong}]{fabbri-etal-2022-qafacteval}
Alexander Fabbri, Chien-Sheng Wu, Wenhao Liu, and Caiming Xiong. 2022.
\newblock \href {https://doi.org/10.18653/v1/2022.naacl-main.187} {{QAF}act{E}val: Improved {QA}-based factual consistency evaluation for summarization}.
\newblock In \emph{Proceedings of the 2022 Conference of the North American Chapter of the Association for Computational Linguistics: Human Language Technologies}, pages 2587--2601, Seattle, United States. Association for Computational Linguistics.

\bibitem[{Gehrmann et~al.(2023)Gehrmann, Clark, and Sellam}]{10.1613/jair.1.13715}
Sebastian Gehrmann, Elizabeth Clark, and Thibault Sellam. 2023.
\newblock \href {https://doi.org/10.1613/jair.1.13715} {{Repairing the Cracked Foundation: A Survey of Obstacles in Evaluation Practices for Generated Text}}.
\newblock \emph{Journal of Artificial Intelligence Research}, 77:103--166.

\bibitem[{Grusky et~al.(2018)Grusky, Naaman, and Artzi}]{N18-1065}
Max Grusky, Mor Naaman, and Yoav Artzi. 2018.
\newblock \href {http://aclweb.org/anthology/N18-1065} {Newsroom: A dataset of 1.3 million summaries with diverse extractive strategies}.
\newblock In \emph{Proceedings of the 2018 Conference of the North American Chapter of the Association for Computational Linguistics: Human Language Technologies}, pages 708--719, New Orleans, Louisiana. Association for Computational Linguistics.

\bibitem[{Hermann et~al.(2015)Hermann, Ko\v{c}isk\'{y}, Grefenstette, Espeholt, Kay, Suleyman, and Blunsom}]{10.5555/2969239.2969428}
Karl~Moritz Hermann, Tom\'{a}\v{s} Ko\v{c}isk\'{y}, Edward Grefenstette, Lasse Espeholt, Will Kay, Mustafa Suleyman, and Phil Blunsom. 2015.
\newblock Teaching machines to read and comprehend.
\newblock In \emph{Proceedings of the 28th International Conference on Neural Information Processing Systems - Volume 1}, NIPS'15, page 1693–1701, Cambridge, MA, USA. MIT Press.

\bibitem[{Laban et~al.(2022{\natexlab{a}})Laban, Schnabel, Bennett, and Hearst}]{laban2022summac}
Philippe Laban, Tobias Schnabel, Paul~N Bennett, and Marti~A Hearst. 2022{\natexlab{a}}.
\newblock Summac: Re-visiting nli-based models for inconsistency detection in summarization.
\newblock \emph{Transactions of the Association for Computational Linguistics}, 10:163--177.

\bibitem[{Laban et~al.(2022{\natexlab{b}})Laban, Schnabel, Bennett, and Hearst}]{10.1162/tacl_a_00453}
Philippe Laban, Tobias Schnabel, Paul~N. Bennett, and Marti~A. Hearst. 2022{\natexlab{b}}.
\newblock \href {https://doi.org/10.1162/tacl_a_00453} {{SummaC: Re-Visiting NLI-based Models for Inconsistency Detection in Summarization}}.
\newblock \emph{Transactions of the Association for Computational Linguistics}, 10:163--177.

\bibitem[{Lin(2004)}]{lin-2004-rouge}
Chin-Yew Lin. 2004.
\newblock \href {https://aclanthology.org/W04-1013} {{ROUGE}: A package for automatic evaluation of summaries}.
\newblock In \emph{Text Summarization Branches Out}, pages 74--81, Barcelona, Spain. Association for Computational Linguistics.

\bibitem[{Liu et~al.(2022)Liu, Fabbri, Liu, Zhao, Nan, Han, Han, Joty, Wu, Xiong, and Radev}]{10.48550/arxiv.2212.07981}
Yixin Liu, Alexander~R Fabbri, Pengfei Liu, Yilun Zhao, Linyong Nan, Ruilin Han, Simeng Han, Shafiq Joty, Chien-Sheng Wu, Caiming Xiong, and Dragomir Radev. 2022.
\newblock \href {https://doi.org/10.48550/arxiv.2212.07981} {{Revisiting the Gold Standard: Grounding Summarization Evaluation with Robust Human Evaluation}}.
\newblock \emph{arXiv}.

\bibitem[{Narayan et~al.(2018)Narayan, Cohen, and Lapata}]{Narayan2018DontGM}
Shashi Narayan, Shay~B. Cohen, and Mirella Lapata. 2018.
\newblock Don't give me the details, just the summary! topic-aware convolutional neural networks for extreme summarization.
\newblock \emph{ArXiv}, abs/1808.08745.

\bibitem[{Nenkova and Louis(2008)}]{nenkova-louis-2008-summarize}
Ani Nenkova and Annie Louis. 2008.
\newblock \href {https://aclanthology.org/P08-1094} {Can you summarize this? identifying correlates of input difficulty for multi-document summarization}.
\newblock In \emph{Proceedings of ACL-08: HLT}, pages 825--833, Columbus, Ohio. Association for Computational Linguistics.

\bibitem[{Novikova et~al.(2018)Novikova, Dusek, and Rieser}]{DBLP:conf/naacl/NovikovaDR18}
Jekaterina Novikova, Ondrej Dusek, and Verena Rieser. 2018.
\newblock \href {https://doi.org/10.18653/v1/n18-2012} {Rankme: Reliable human ratings for natural language generation}.
\newblock In \emph{Proceedings of the 2018 Conference of the North American Chapter of the Association for Computational Linguistics: Human Language Technologies, NAACL-HLT, New Orleans, Louisiana, USA, June 1-6, 2018, Volume 2 (Short Papers)}, pages 72--78. Association for Computational Linguistics.

\bibitem[{Papineni et~al.(2002)Papineni, Roukos, Ward, and Zhu}]{papineni-etal-2002-bleu}
Kishore Papineni, Salim Roukos, Todd Ward, and Wei-Jing Zhu. 2002.
\newblock \href {https://doi.org/10.3115/1073083.1073135} {{B}leu: a method for automatic evaluation of machine translation}.
\newblock In \emph{Proceedings of the 40th Annual Meeting of the Association for Computational Linguistics}, pages 311--318, Philadelphia, Pennsylvania, USA. Association for Computational Linguistics.

\bibitem[{Rankel et~al.(2011)Rankel, Conroy, Slud, and O{'}Leary}]{rankel-etal-2011-ranking}
Peter Rankel, John Conroy, Eric Slud, and Dianne O{'}Leary. 2011.
\newblock \href {https://aclanthology.org/D11-1043} {Ranking human and machine summarization systems}.
\newblock In \emph{Proceedings of the 2011 Conference on Empirical Methods in Natural Language Processing}, pages 467--473, Edinburgh, Scotland, UK. Association for Computational Linguistics.

\bibitem[{See et~al.(2017)See, Liu, and Manning}]{see-etal-2017-get}
Abigail See, Peter~J. Liu, and Christopher~D. Manning. 2017.
\newblock \href {https://doi.org/10.18653/v1/P17-1099} {Get to the point: Summarization with pointer-generator networks}.
\newblock In \emph{Proceedings of the 55th Annual Meeting of the Association for Computational Linguistics (Volume 1: Long Papers)}, pages 1073--1083, Vancouver, Canada. Association for Computational Linguistics.

\bibitem[{Zhang* et~al.(2020)Zhang*, Kishore*, Wu*, Weinberger, and Artzi}]{bert-score}
Tianyi Zhang*, Varsha Kishore*, Felix Wu*, Kilian~Q. Weinberger, and Yoav Artzi. 2020.
\newblock \href {https://openreview.net/forum?id=SkeHuCVFDr} {Bertscore: Evaluating text generation with bert}.
\newblock In \emph{International Conference on Learning Representations}.

\bibitem[{Zhong et~al.(2022)Zhong, Liu, Yin, Mao, Jiao, Liu, Zhu, Ji, and Han}]{zhong-etal-2022-towards}
Ming Zhong, Yang Liu, Da~Yin, Yuning Mao, Yizhu Jiao, Pengfei Liu, Chenguang Zhu, Heng Ji, and Jiawei Han. 2022.
\newblock \href {https://aclanthology.org/2022.emnlp-main.131} {Towards a unified multi-dimensional evaluator for text generation}.
\newblock In \emph{Proceedings of the 2022 Conference on Empirical Methods in Natural Language Processing}, pages 2023--2038, Abu Dhabi, United Arab Emirates. Association for Computational Linguistics.

\end{thebibliography}

\clearpage

\section*{Appendix}
\label{sec:appendix}
\renewcommand{\thesubsection}{\Alph{subsection}}

\subsection{Summarization Prompt Details}
\label{appendix:prompts}
For the summarization prompts, we use prompts and input structures that the models have been trained on. Table \ref{tab:prompts} shows the input for each model, where \texttt{[TEXT]} is replaced with the article to be summarized.
\begin{table}[]
\scalebox{0.6}{
\begin{tabular}{ll}
\toprule
Model    & Prompt                                                                                                                       \\ \midrule
FlanT5   & \texttt{[TEXT]}\textbackslash{}nWhat is a one-paragraph summary of the above article?                                                                                                                         \\ \midrule
StableLM & \begin{tabular}[c]{@{}l@{}}\textless{}|SYSTEM|\textgreater{}\# StableLM Tuned (Alpha version)\\ - StableLM is a helpful and harmless open-source AI language \\  model developed by StabilityAI.\\ - StableLM is able to facilitate human communication by \\ providing a summary of a given text.\\ - StableLM is able to provide summaries that are useful and \\ relevant to the given text.\\ \textless{}|USER|\textgreater \ \texttt{[TEXT]}. \\ Summarize the given piece of text.\\ \textless{}|ASSISTANT|\textgreater{}\end{tabular} \\ \midrule
GPT-4    & \begin{tabular}[c]{@{}l@{}}"role": "user", \\ "content": ''\texttt{[TEXT]} \textbackslash{}n\textbackslash{}n Summarize the above text. \textbackslash{}n\textbackslash{}n"\end{tabular}                        
\end{tabular}
}
\caption{Input and prompt structure for each summarization model. \texttt{[TEXT]} is replaced with the article to be summarized.}
\label{tab:prompts}
\end{table}

\subsection{BLEU and SummaC-ZS}
\label{appendix:addl_violin_plots}
\begin{figure*}
    \centering
    \includegraphics[width=\textwidth]{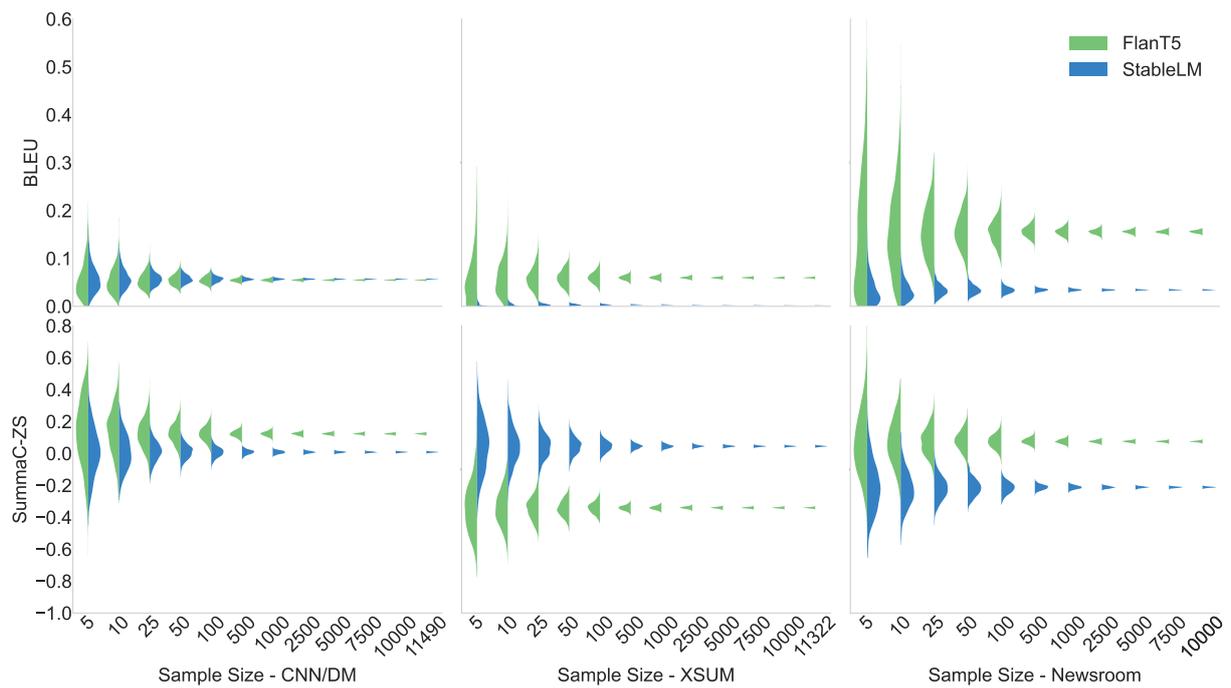}
    \caption{Distributions of averaged BLEU and SummaC-ZS scores across 1000 re-samples for CNN/DM, XSUM, and Newsroom.} 
    \label{fig:appendix_full_plots}
\end{figure*}

Figure \ref{fig:appendix_full_plots} shows the distributions of averaged BLEU and SummaC-ZS scores over all three datasets. BLEU scores have trouble capturing meaningful scores across longer inputs as seen with StableLM. SummaC-ZS uses NLI-models to score sentence-level information -- similar to ROUGE-1 and BERTScore, we can start differentiating models earlier than the full sample size. 

\subsection{Human Evaluation Win Rates and Sample Sizes: XSUM and Newsroom}
\label{appendix:human_eval_win_rates}
\begin{figure*}
    \centering
    \includegraphics[width=\textwidth]{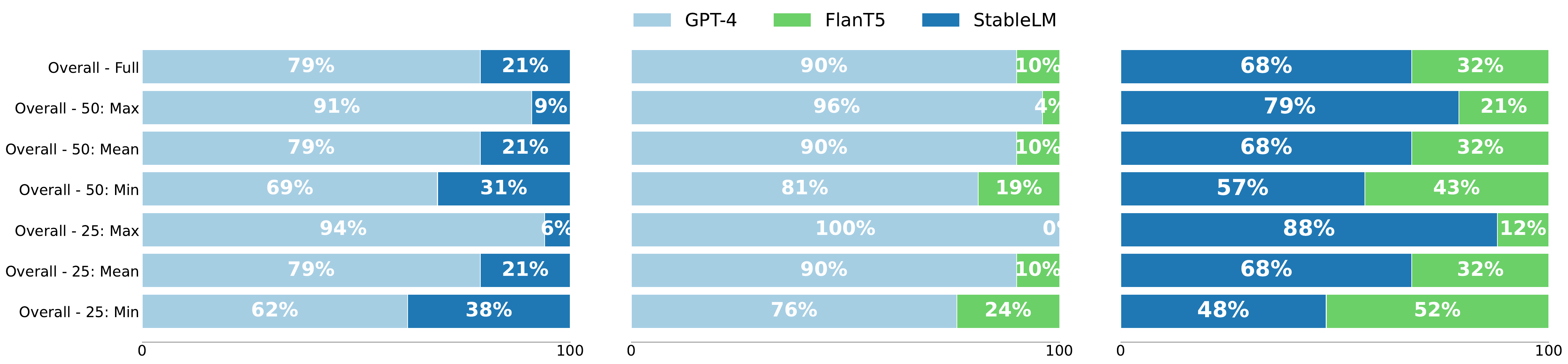}
    \caption{Win rates aggregated by annotators (XSUM).}
    \label{fig:appendix_win_rates_xsum}
\end{figure*}

\begin{figure*}
    \centering
    \includegraphics[width=\textwidth]{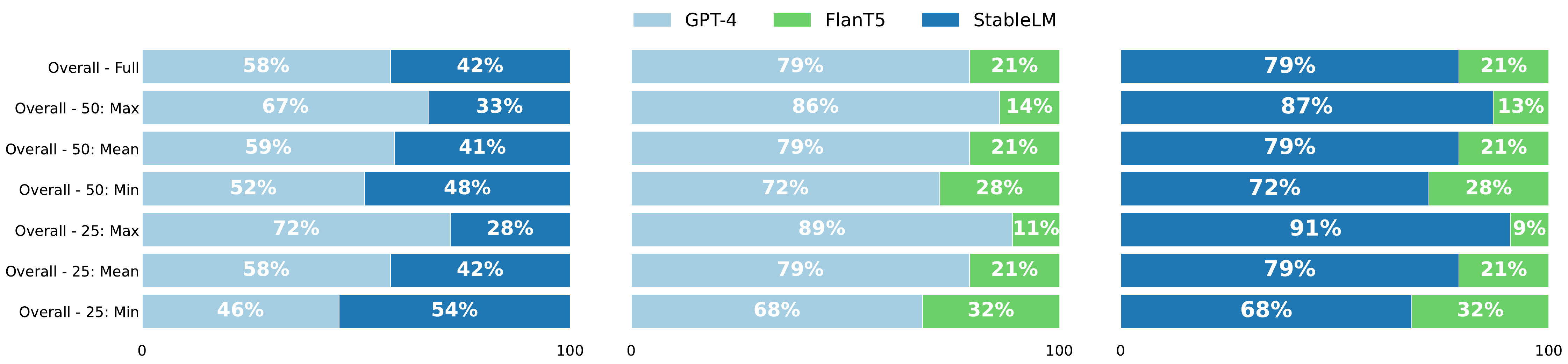}
    \caption{Win rates aggregated by annotators (Newsroom).}
    \label{fig:appendix_win_rates_news}
\end{figure*}

We provide the aggregated win rates across annotators for XSUM (Figure \ref{fig:appendix_win_rates_xsum}) and Newsroom (Figure \ref{fig:appendix_win_rates_news}). Both datasets show the same trend as in Figure \ref{fig:win_rates_cnn}, where the win rate pair ranking is preserved in the minimum, maximum, and average win rates across 1000 trials. This holds across sample sizes of 50, but not in \textit{all} cases with sample size of 25. 


\subsection{Human Evaluation Win Rates and Tasks: XSUM and Newsroom}
\begin{figure*}
    \centering
    \includegraphics[width=\textwidth]{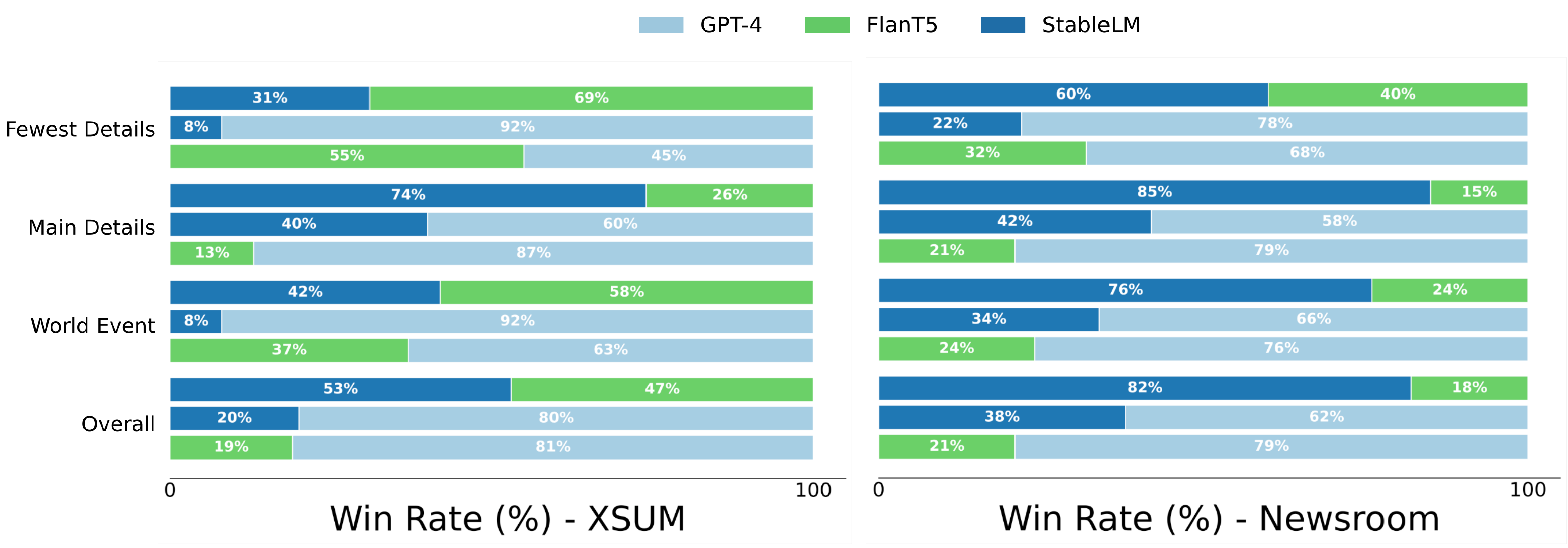}
    \caption{ Aggregated annotator win rates across all metrics over the XSUM and Newsroom datasets. }
    \label{fig:win_rates_xsum_news_all_metrics}
\end{figure*}

Similar to Figure \ref{fig:win_rates_cnn_all_metrics}, we show the win rates across different tasks for XSUM and Newsroom in Figure \ref{fig:win_rates_xsum_news_all_metrics}. These results support the finding that preference changes between downstream scenarios.

\subsection{Annotator Agreement on Text Quality and Factuality}
\label{appendix:text_quality_factuality}
 For CNN/DM we report the agreement scores over factuality and text quality questions that we collect in our surveys in Table \ref{table:quality_agreement}. We expect the agreement scores for factuality to be much higher; it is possible that this is an indicator for different tolerance for minor errors (e.g., vague wording) or may be indicative of the cognitive load involved in judging factuality. Similarly for text quality, the threshold for artifacts or other issues may differ between annotators. 

\begin{table}[]
\scalebox{1}{
\begin{tabular}{@{}ccc@{}}
\toprule
\multicolumn{3}{l}{CNN/DM}                                        \\ \midrule
Annotators & Factuality $\kappa$ & Text Quality $\kappa$  \\ \midrule
1, 2       & 0.522                  & 0.053                    \\
1, 3       & 0.249                  & 0.539                    \\
2, 3       & 0.133                  & -0.081                   \\ \midrule
                                \\
\end{tabular}
}
\caption{Agreement scores, Cohen's kappa.}
\label{table:quality_agreement}
\end{table}

\subsection{Annotation Details}
\label{appendix:annotation_details}
\paragraph{Costs}
We hired seven professional proofreaders from Upwork, who were each recruited to read 100 articles and rank 3 summaries per article.
We paid each annotator a flat fee of \$325 to evaluate the summaries
When asked for a time estimate after they completed, responses ranged between 10 and 13 hours to complete the study, meaning annotators were compensated at roughly \$25-\$30 per hour.
The annotators typically completed the work over one to three days.

\begin{figure*}
    \centering
     \begin{subfigure}[t]{0.45\textwidth}
         \centering
         \includegraphics[width=\textwidth]{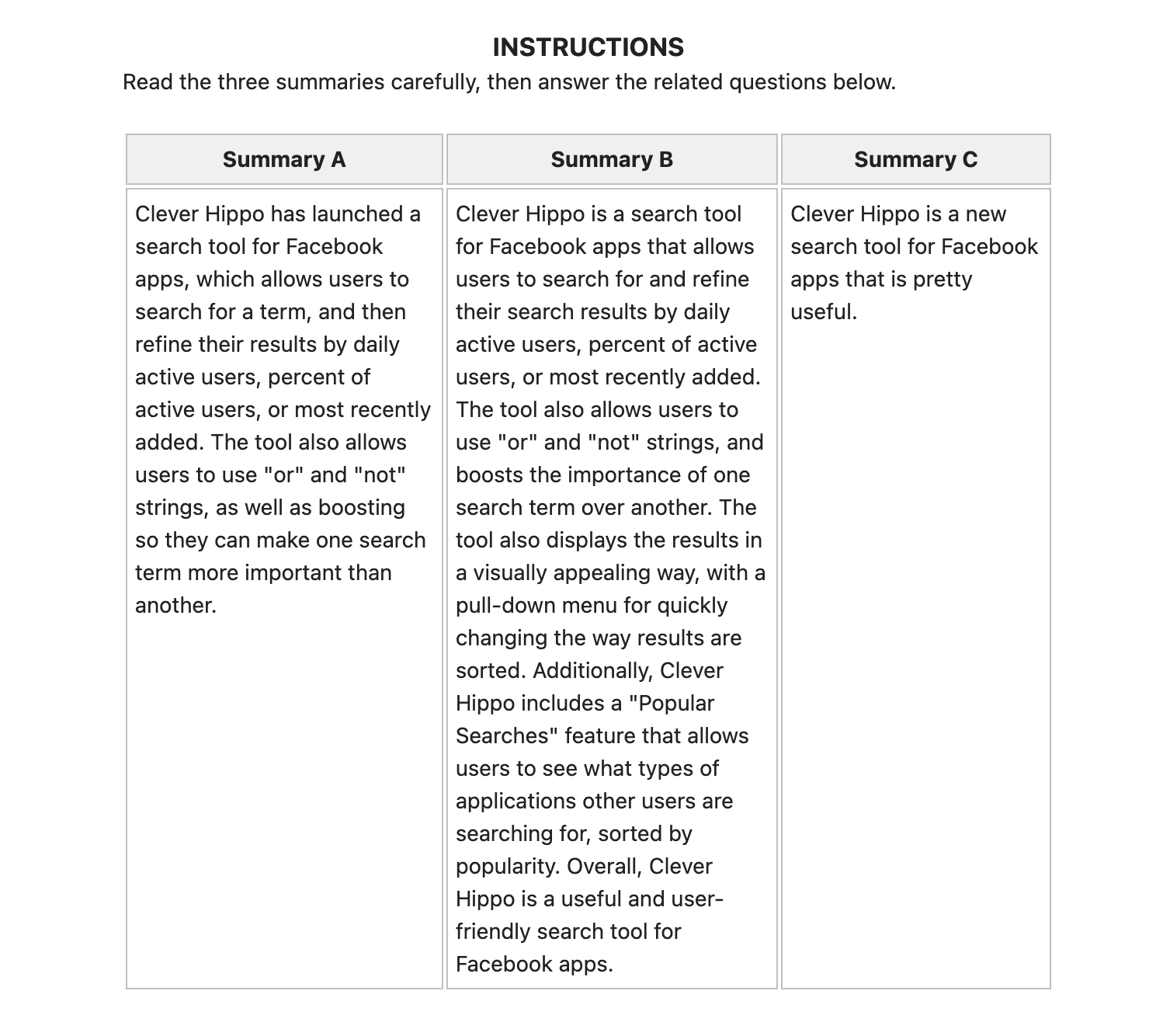}
         \caption{Summaries, as presented to the annotators.}
         \label{subfig:interface_summaries}
     \end{subfigure}
     \hfill
     \begin{subfigure}[t]{0.45\textwidth}
         \centering
         \includegraphics[width=\textwidth]{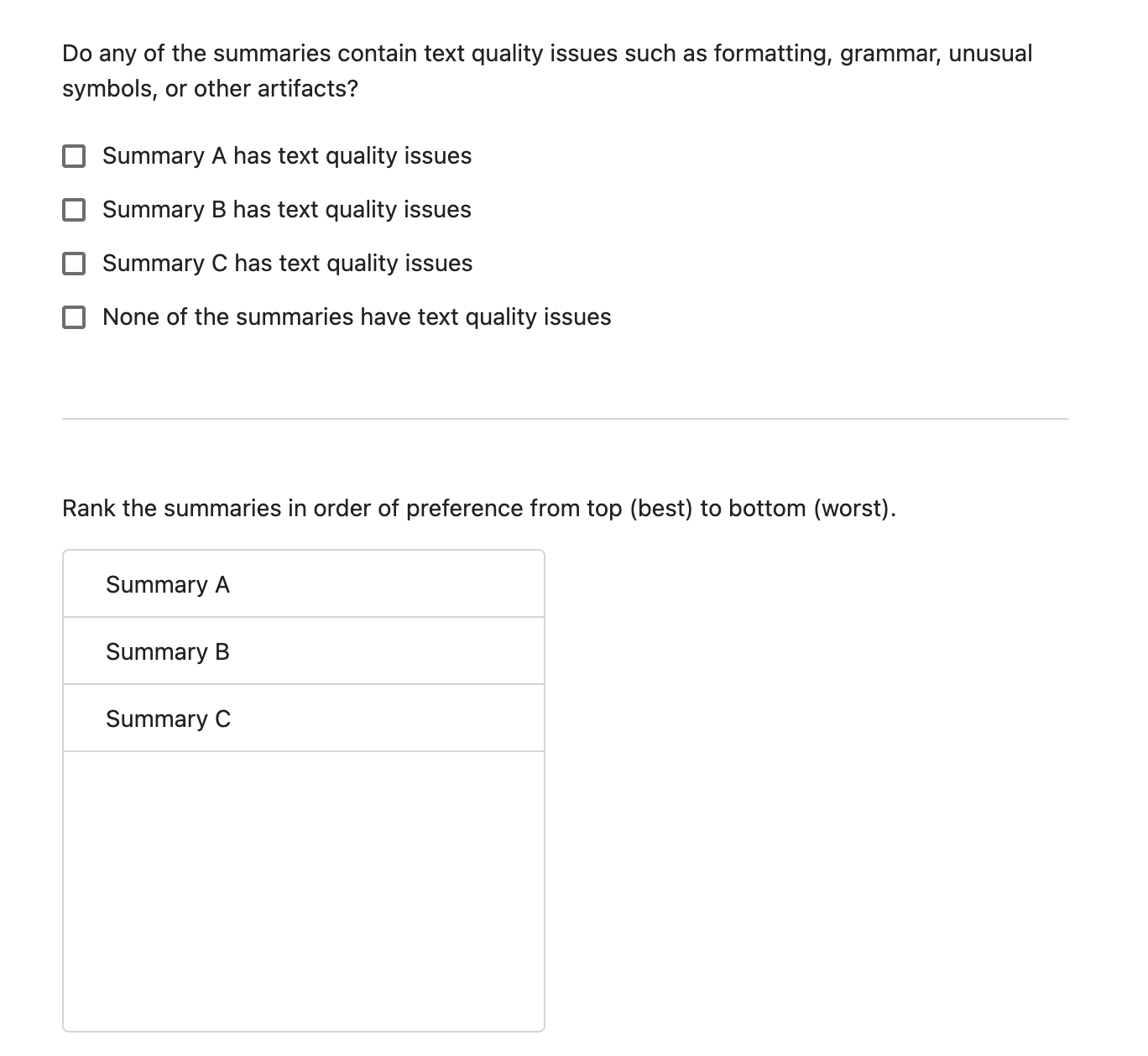}
         \caption{Text quality issues and the ranking interface for the summaries. Each box with the summary label can be dragged-and-dropped into any order.}
         \label{subfig:interface_ranking}
     \end{subfigure}
     \hfill \\
     \begin{subfigure}[t]{0.45\textwidth}
         \centering
         \includegraphics[width=\textwidth]{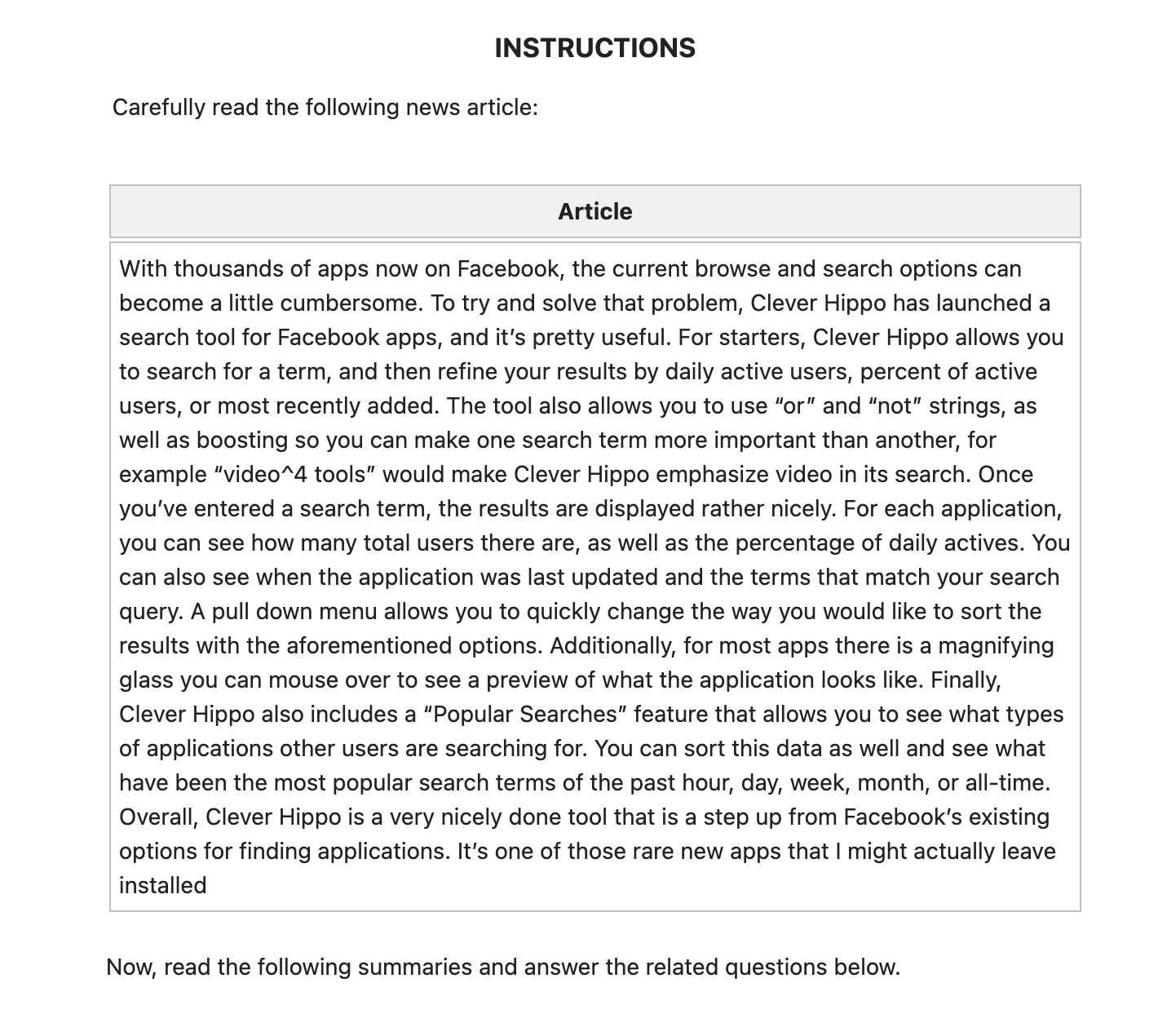}
         \caption{Article, as presented to the annotator.}
         \label{subfig:interface_article}
     \end{subfigure}
     \hfill
     \begin{subfigure}[t]{0.45\textwidth}
         \centering
         \includegraphics[width=\textwidth]{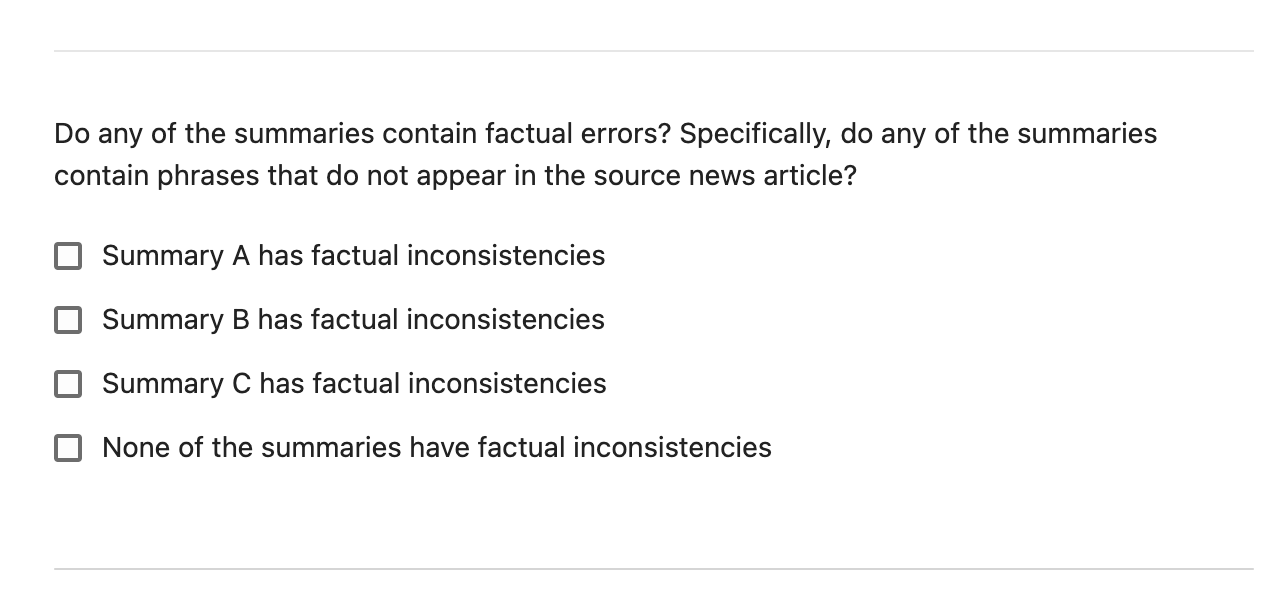}
         \caption{Factuality questions asked about each summary.}
         \label{subfig:interface_factual}
     \end{subfigure}
        \caption{The annotation interface. For each article, annotation happens across two pages. The first page contains the summaries (\ref{subfig:interface_summaries}) and rankings (\ref{subfig:interface_ranking}), and the second page contains the article (\ref{subfig:interface_article}) and factuality questions (\ref{subfig:interface_factual}).}
        \label{fig:interface}
\end{figure*}

\paragraph{Annotation Platform}
We presented the annotators with a custom interface for ranking the summaries and answering questions, shown in Figure~\ref{fig:interface}.
Annotators were encouraged to take extended breaks during annotation to reduce task fatigue. 
\end{document}